# Multi-pretrained Deep Neural Network

Zhen Hu[1,2,a,*], Zhuyin Xue[1], Tong Cui[1], Shiqiang Zong[1], Chenglong He[1]

[1]Science and Technology on Information Systems Engineering Laboratory,

Nanjing Research Institute of Electronic and Engineering,

Nanjing China, 210007

[2]CETC Research Institute of Smart City,

Shenzhen China, 518000

[a]49859211@qq.com

*Corresponding author

**Abstract.** Pretraining is widely used in deep neutral network and one of the most famous pretraining models is Deep Belief Network (DBN). The optimization formulas are different during the pretraining process for different pretraining models. In this paper, we pretrained deep neutral network by different pretraining models and hence investigated the difference between DBN and Stacked Denoising Autoencoder (SDA) when used as pretraining model. The experimental results show that DBN get a better initial model. However the model converges to a relatively worse model after the finetuning process. Yet after pretrained by SDA for the second time the model converges to a better model if finetuned.

**Introduction**

Neural Network has long been a widely used model in machine learning. In 1982 Hopfield proposed the Hopfield Network[8] and proved that neural network can be used to simulate the XOR function. In 1986, Hinton et. al proposed the Backpropagation algorithm(BP algorithm) to train multiple-layer neural network[16] and henceforth neural network has become a widely used model in machine learning areas such as image processing[5], control[9] and optimization[14].

However, the training process of neural network is a non-convex problem while the BP algorithm is intrinsically a gradient-descent algorithm which is guaranteed to converge to a global optima when solving convex problems[1]. This very property makes the BP algorithm converges to a local optima which is severely dependent to the Initial state of the network. In real cases, researchers usually randomly choose different Initial states before training the network and at last choose the best-performed model. This trick is very inefficient and even unbearable when training large-scale network. Some researchers proposed improved algorithms such as simulated annealing[4], Genetic algorithm[19] to reduce the training epoches. However the progress was not so expecting. What's more, as the powerful expressive ability of neural network, a network in the global optimal state maybe severely over-fitting[17], and even worse than a network in the local optimal state. As the fast development fast-learnable model such as Support Vector Machine[6], researchers were more and more considerable about the training data[3]. As the scale of training data rises, more complicated neural network were proposed and the harder the models were trained by BP algorithm.

To relief the training burden, Lecun et. al introduced the parameter bonding strategy into the training process of neural network[12]. In their model, they bonded the parameter to suit some kind of prior knowledge to restrict the expression power of neural network. And by doing so, the number of local optima was reduced which made the training of deep neural network possible. In the parameter bonding training process, the global optima is not important any more. In fact, by bonding the parameters, the global optima is un-reachable. The work of Lecun et.al inspired researchers to divide the training process into two stage. One for feature extracting which was named as feature learning and one for classification. Lecun et. al proposed the model named as LeNet5[12] which was a 8-layer neural network. The parameter bonding was conducted by replacing product of weight matrix and input vector by convolution of kernel and input vector. The prior was the shifting invariant of images. The proposed model solved the MNIST problem well.

Another strategy to solve the large-scale machine learning problem was focus on the initial state. In 2006, Hinton et. al proposed the pre-training strategy[7]. The pre-training process aims at seeking a well-performed initial state of the neural network which was fine-tuned by the BP algorithm. The fine-tuning process was conducted only once. Since the pre-training process was layer-wised, the computation complexity increased linearly an the number of layers raised. The pre-training process proposed by Hinton et. al took the conjoint layers as Restricted Boltzmann Machine (RBM) and pre-trained the layers by maximizing the likelihood function of RBM. Some researchers used models other than RBM to pre-train the neural network. Honglak Lee et. al proposed the convolutional RBM[13] while Larochelle et. al proposed the auto-encoder (AE) [11]. In convolutional RBM, the computation rule between layers is convolution other than production and in AE the optimization problem is to minimize the reconstruction error other than maximizing the likelihood function. Some researchers improved AE, such as De-noising Auto-encoder (DAE)[18], Contractive Autoencoder (CAE)[15] and so on.

In our work, we proposed a new pre-training process to get a better initial state. We want to combine the advantage of different models to pre-train the network as much as possible. We proposed the Multi Pre-trained Deep Neural Network (MPDNN). In our model, we pre-trained the network by RBM and DAE multi-times and the experiment results show that the network performed best when pre-trained by RBM and then pre-trained by DA.

The paper was arranged as follow. In section 2, we introduced our proposed model. we reported our experimental results in section 3 we also tested different pre-training strategies in this section. In section 4 we concluded our work.

**Model**

In this section, we introduced our proposed model. Since our model was based on RBM and DAE, we introduced these two models first.

**Restricted Boltzmann Machine**

In RBM, nodes were divided into visible nodes and hidden nodes. The visible nodes took the original data as input and the hidden nodes were not directly connected to the input. The value of visible and hidden nodes were denoted by v and h respectively.

In RBM, an energy function was defined as
$$E(v, h) = h^T W v + b^T h + c^T v. \tag{1}$$
where W was the weight matrix, b and c were the visible and hidden biases. The likelihood function was defined as
$$L(v) = \log \sum_h P(v, h) = \log \frac{\sum_h e^{-E(v,h)}}{\sum_v \sum_h e^{-E(v,h)}}. \tag{2}$$
It's easy to see that values of hidden nodes were conditionally independent on condition of visible nodes and values of visible nodes were conditionally independent on condition of hidden nodes[10]. So equation (2) can be optimized by gibbs sampling[2]. Hinton proposed that the gibbs sampling process can be only run for once to get a well enough initial state[7].

**De-noising Auto-encoder**

DAE was a variant of AE. In AE, values of hidden nodes were calculated by the following equation,
$$h = \text{sigm}(Wv + b), \tag{3}$$
where sigm denoted the non-linear sigmoid function. The reconstruction process was fulfilled by
$$\hat{v} = \text{sigm}(W^T h + c), \tag{4}$$
The optimization problem for AE was defined as
$$\min_{W,b,c} \|v - \hat{v}\|_2. \tag{5}$$

Vincent et. al argued that the reconstruction process was so simple that AE was risking to be over-fitting. They introduced the corruption strategy in which the input v was corrupted by randomly set some elements to 0 first to get the corrupted input $v$. After encoded and decoded as AE, the reconstructed input was compared with the clean input $v$. Their experiment showed that by corrupting the input, the network was more general.

**Multi-pretrained Deep Neural Network**

Our proposed model, MPDNN, was based on RBM and DAE. We use these models to pre-train our network twice and the experimental results showed the difference.

The optimization problem for RBM was a likelihood maximization problem and that for DAE was an error minimization problem. So DAE was more loyal to the original input and RBM can be varied. These difference can not be erased by the fine-tuning process.

**Experiment**

In this section, we report our experiment results.

**Experiment Setup**

Our proposed model, MPDNN, was based on RBM and DAE. We use these models to pre-train our network twice and the experimental results showed the difference.

The optimization problem for RBM was a likelihood maximization problem and that for DAE was an error minimization problem. So DAE was more loyal to the original input and RBM can be varied. These difference can not be erased by the fine-tuning process.

In our paper, we considered a widely used benchmark problem, the MNIST handwritten digits recognition problem. In this problem, we picked 50,000 pictures as the training set, 10,000 pictures as the validation set and 10,000 as the test set.

We proposed a 4-layer network to recognize the digits. The number of hidden nodes in all 3 hidden layers was 1,000. In our experiment we implemented four models,
 -MPDNN-SS, the network was pre-trained by DAE twice,
 -MPDNN-SD, the network was pre-trained by DAE first and then pre-trained by RBM,
 -MPDNN-DS, the network was pre-trained by RBM first and then DAE
 -MPDNN-DD, the network was pre-trained by RBM twice.

All these models were implemented by Theano and run on NVIDIA C2075. We run every model for 50 times and table 1 showed the result.

**Experiment Results**

The results showed that, MPDNN-DD was dull to the initial state as the variance for all 50 runs was smallest. MPDNN-DS performed best on accuracy among all the models. Once pretrained by DAE, no matter followed by DAE or RBM, the network performed worse, with either larger error rate or larger variance.

Since the pre-training process was layer-wise, MPDNN-DD was equivalent to RBM with more pre-training epoches, and MPDNN-SS was equivalent to DAE with more pre-training epoches.

Table 1. Performance of four models in our paper

| Modelt | Error rate on validation se | Error rate on test set |
|---|---|---|
| MPDNN-SS | 1.85%±0.06% | 1.99%±0.08% |
| MPDNN-SD | 1.96%±0.09% | 2.12%±0.09% |
| MPDNN-DS | 1.82%±0.05% | 1.93%±0.08% |
| MPDNN-DD | 1.84%±0.04% | 1.98%±0.06% |

Figure. 1 showed the curve of error rate along with fine-tuning iteration. We find that MPDNN-DX (X=D,S) performed better than MPDNN-SX (X=D,S) from the beginning of fine-tuning till the end. And as the fine-tuning iteration raised MPDNN-DS became better than MPDNN-SS, even though MPDNN-DS was worse than MPDNN-SS in the beginning. The phenomenon indicated that RBM was a better model to find a relatively good initial state while the model converged to a poor-performed local minima when fine-tuned. However, once pre-trained by DAE for the second time, the model performed better.

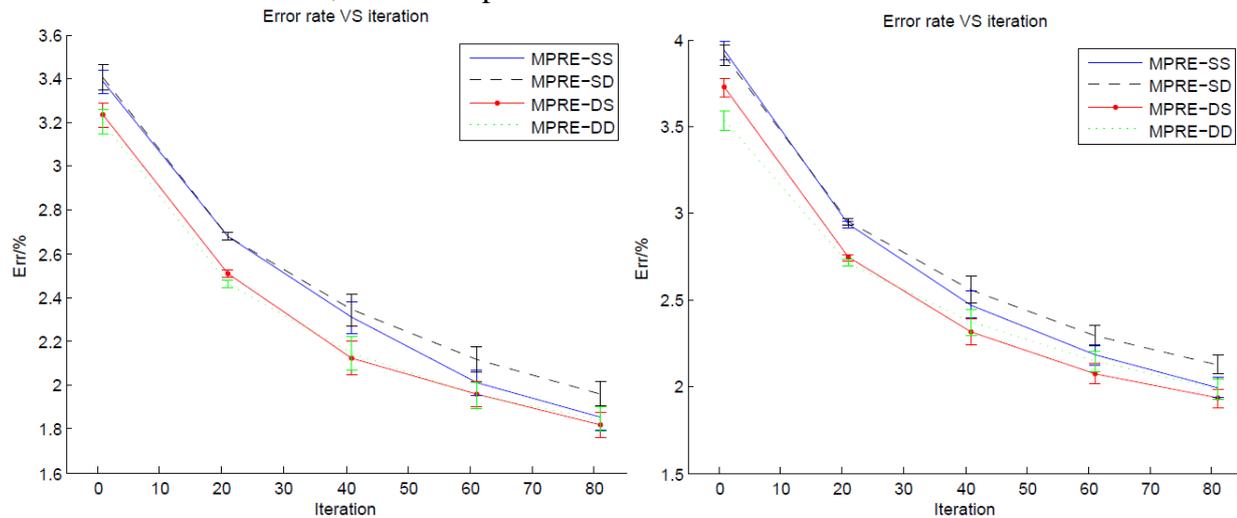

Figure 1.a Error rate on validation set   Figure 1.b Error rate on test set
Figure 1 Performance on validation and test sets.

**Summary**

In this work, we used different models to pre-train a deep neural network, and eventually got a better performed model. The experimental results showed that RBM was good at finding a better initial state for the neural network while this initial state may get stuck when fine-tuning the model. By pre-training the model using DAE for the second time, the model may performed worse than the model pre-trained by RBM at the beginning of fine-tuning, and eventually surpass the RBM pre-trained model as the fine-tuning process goes.

In our work, we only compared two models, that is RBM and DAE. We will test more models in the near future to find the advantage of different pre-training models. By doing this, we can choose the most suitable pre-training model when we train our network.

**Acknowledgement**

The research work was supported by National Natural Science Foundation of China under Grant No. 61402426, Training Program of the Major Research Plan of the National Natural Science Foundation of China (No. 91546103)  and partially supported by Collaborative Innovation Center of Social Safety Science and Technology.

**References**

[1]  Boyd, S., Vandenberghe, L.: Convex optimization. Cambridge university press (2004)
[2]  Carter, C.K., Kohn, R.: On gibbs sampling for state space models. Biometrika 81(3), 541–553 (1994)
[3]  Church, K.: A pendulum swung too far. Linguistic Issues in Language Technology 6(5), 1–27 (2011)
[4]  Goffe, W.L., Ferrier, G.D., Rogers, J.: Global optimization of statistical functions with simulated annealing. Journal of econometrics 60(1-2), 65–99 (1994)
[5]  Hagan, M.T., Demuth, H.B., Beale, M.H., De Jes ús, O.: Neural network design, vol. 20. PWS publishing company Boston (1996)

[6] Hearst, M.A., Dumais, S.T., Osman, E., Platt, J., Scholkopf, B.: Support vector machines. Intelligent Systems and their Applications, IEEE 13(4), 18–28 (1998)

[7] Hinton, G.E., Osindero, S., Teh, Y.W.: A fast learning algorithm for deep belief nets. Neural computation 18(7), 1527–1554 (2006)

[8] Hopfield, J.J.: Neural networks and physical systems with emergent collective computational abilities. Proceedings of the national academy of sciences 79(8), 2554–2558 (1982)

[9] Hunt, K.J., Sbarbaro, D., Żbikowski, R., Gawthrop, P.J.: Neural networks for control systemsa survey. Automatica 28(6), 1083–1112 (1992)

[10] Koller, D., Friedman, N.: Probabilistic graphical models: principles and techniques. MIT press (2009)

[11] Larochelle, H., Erhan, D., Courville, A., Bergstra, J., Bengio, Y.: An empirical evaluation of deep architectures on problems with many factors of variation. In Proceedings of the 24th international conference on Machine learning. pp. 473–480. ACM (2007)

[12] LeCun, Y., Bottou, L., Bengio, Y., Haffner, P.: Gradient-based learning applied to document recognition. Proceedings of the IEEE 86(11), 2278–2324 (1998)

[13] Lee, H., Grosse, R., Ranganath, R., Ng, A.Y.: Convolutional deep belief networks for scalable unsupervised learning of hierarchical representations. In Proceedings of the 26th Annual International Conference on Machine Learning. pp. 609–616. ACM (2009)

[14] Potvin, J.Y.: State-of-the-art surveythe traveling salesman problem: A neural network perspective. ORSA Journal on Computing 5(4), 328–348 (1993)

[15] Rifai, S., Vincent, P., Muller, X., Glorot, X., Bengio, Y.: Contractive auto-encoders: Explicit invariance during feature extraction. In Proceedings of the 28th international conference on machine learning (ICML-11). pp. 833–840 (2011)

[16] Rumelhart, D., Hinton, G., Williams, R.: Learning internal representation by back propagation. Parallel distributed processing: exploration in the microstructure of cognition 1 (1986)

[17] Srinivas, M., Patnaik, L.M.: Adaptive probabilities of crossover and mutation in genetic algorithms. Systems, Man and Cybernetics, IEEE Transactions on 24(4), 656–667 (1994)

[18] Vincent, P., Larochelle, H., Lajoie, I., Bengio, Y., Manzagol, P.A.: Stacked denoising autoencoders: Learning useful representations in a deep network with a local denoising criterion. The Journal of Machine Learning Research 11, 3371–3408 (2010)

[19] Whitley, D., Starkweather, T., Bogart, C.: Genetic algorithms and neural networks: Optimizing connections and connectivity. Parallel computing 14(3), 347–361 (1990)